\documentclass[pdflatex,sn-basic]{sn-jnl}

\usepackage{anyfontsize}
\usepackage{changepage,threeparttable}
\usepackage{tabularx}
\usepackage{commath}
\usepackage{makecell}
\jyear{2023}%

\theoremstyle{thmstyleone}%
\theoremstyle{thmstyletwo}%

\theoremstyle{thmstylethree}%

\begin{document}

\title[Fake News Detection using Dual BERT Deep Neural Networks]{Fake News Detection using Dual BERT Deep Neural Networks}


\author*[1]{\fnm{Mahmood} \sur{Farokhian}}\email{farokhian@gmail.com}

\author[1]{\fnm{Vahid} \sur{Rafe}}\email{v-rafe@araku.ac.ir}

\author[2]{\fnm{Hadi} \sur{Veisi}}\email{h.veisi@ut.ac.ir}

\affil[1]{\orgdiv{Department of Computer Engineering}, \orgname{Arak University}, \city{Arak}, \country{Iran}}

\affil[2]{\orgdiv{Department of New Science and Technology}, \orgname{University of Tehran}, \city{Tehran}, \country{Iran}}

\abstract{Fake news is a growing challenge for social networks and media. Detection of fake news always has been a problem for many years, but the evolution of social networks and increasing speed of news dissemination in recent years has been considered again. There are several approaches to solving this problem, one of which is to detect fake news based on its text style using deep neural networks. In recent years, transfer learning with transformers is one of the most used forms of deep neural networks for natural language processing. BERT is one of the most promising transformers that outperforms other models in many NLP benchmarks. In this article, we introduce \textbf{\textit{MWPBert}}, which uses two parallel BERT networks to perform veracity detection on full-text news articles. One of the BERT networks encodes news headline, and another encodes news bodies. Since the input length of the BERT network is limited and constant and the news body is usually a long text, we cannot feed the whole text into the BERT. Therefore, using the \textbf{\textit{MaxWorth}} algorithm, we selected the part of the news text that is more valuable for fact-checking, and fed it into the BERT network. Finally, we encode the output of the two BERT networks to an output network to classify the news. The experiment results showed that the proposed model outperformed previous models regarding accuracy and other performance measures.}




\maketitle

\section{Introduction}\label{sec1}

\begin{figure}[t]
\centering
\includegraphics[width=0.9\textwidth]{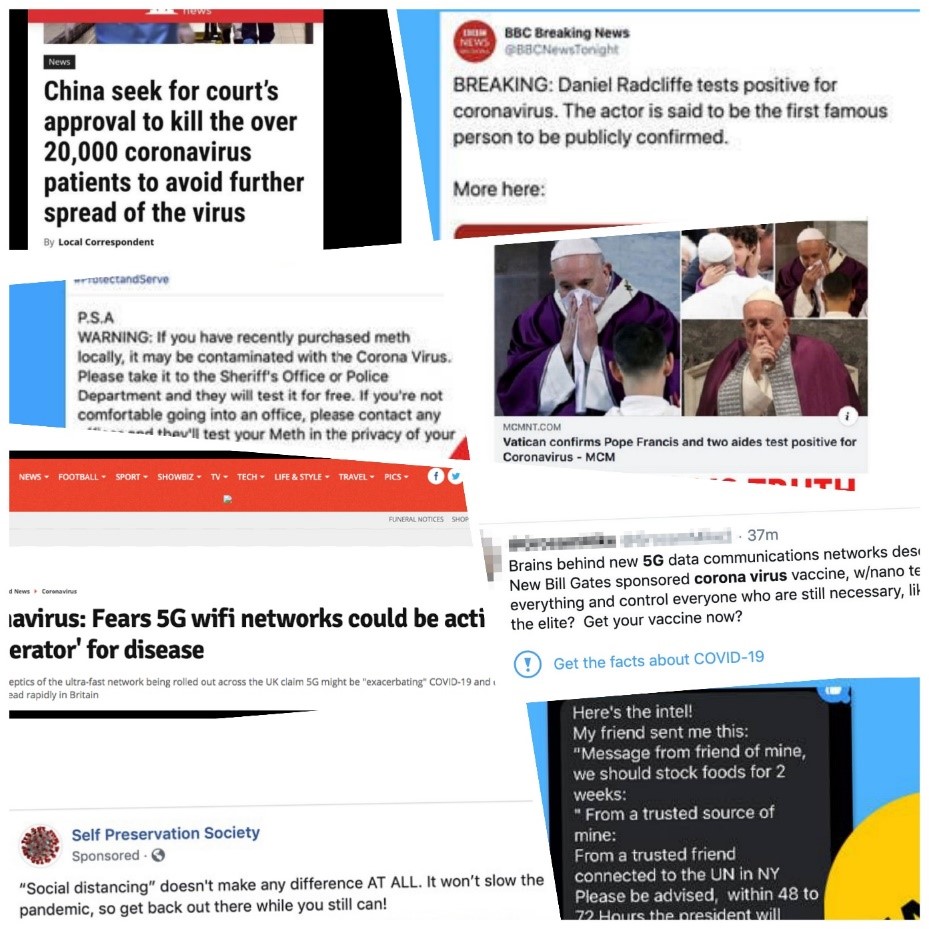}
\caption{Samples of fake news about Covid-19 pandemic}\label{fig1}
\end{figure}

Although the fake news dilemma is not a new concept, in recent years, the emergence of the internet and social networks has changed many of the ideas associated with it and we need to reconsider it. Nowadays, with the widespread use of the internet and mass media, the dissemination of news has become easier and faster than ever before. The significant impact of fake news on the critical events of recent years, namely the US presidential election in 2016, Brexit, and the Covid-19 pandemic, is still not far from memory \cite{Meel2020}. Recent examples of fake news about the Covid-19 epidemic can be seen in ~\ref{fig1}.

Researchers have considered many approaches to solve the problem of fake news. Many methods use the social context of the news to verify its veracity. These methods either examine the pattern of publishing fake and real news on social networks and the different reactions of network users to these two categories or use features such as the author of the news, the source of its publication or the credibility of its returnee users ~\citep{Kwon2013,Wu2015,Zhou2015,Vosoughi2018,Shu2019}. Some other researchers try to detect fake news veracity by evaluating its content. Their works can be divided into two subcategories: the first category uses knowledge-based methods to investigate fake news \citep{Thorne2018,Dagan2009,Pan2018}; In fact, these works treat the news as one or more logical propositions and try to evaluate the degree of compatibility of these propositions with previous knowledge. The second group of these researches uses natural language processing techniques and treats the problem as text classification \cite{Ott2011,Li2017,Abouelenien2017,Perez-Rosas2018,Karimi2019}. In the early works, classification-based approaches tried to extract various features from the news text manually, and these features were used in combination with classical learning methods to determine the veracity of the news. Extracting features from the text, as a complex input, has always been challenging, energy-consuming, and time-consuming for humans.

In recent years, much attention has been paid to deep networks in natural language processing \citep{Monti2019,Bronstein2017,Long2017,Thota2018,Roy2019}. One of the essential advantages of using a deep network for text processing is the automatic feature extraction from the text. In deep neural networks, there is no need to extract and discover features manually from the text; the first layers of the network itself extract the best features from the text. The works that have been done in this area have shown that by a good preprocessing of the input text vectorizing it and feeding to neural networks properly designed for sequence processing, outstanding results can be achieved in natural language processing tasks. Because textual data is sequential, treating it as a time series is best. Classic deep neural networks like Convolutional Neural Network (CNN) do not have this time view of data and do not care about data order \cite{albawi2017understanding}. However, there are neural network architectures that also encode the concept of time into their architecture. Among these deep neural networks are Recurrent Neural Network (RNN), Gated Recurrent Unit (GRU), and Long Short Term Memory (LSTM) \cite{Kowsari2019}. In recent years transformers were introduced as a new neural network architecture, which has delivered outstanding results in NLP tasks \cite{Wolf2020}. The BERT network, which we use in this article, stands for Bidirectional Encoder Representations from Transformers, has achieved one of the best results \cite{Devlin2019}. However, transformers have a critical limitation due to their architecture: the length of the input data must be a constant value. This restriction on news articles, usually not of a certain size and maybe long, can lead to losing important parts of the input data.

In this study, we try to overcome this limitation of the size of the input length of transformers by finding the most useful part of the text that contains valuable information about news trueness- based on the check worthiness of sentences. We provide the news headline and most check worth part of news text into two BERT networks, concatenate both outputs, and feed the result into a linear neural network to categorize the news. The output of this neural network indicates the probability of the truthfulness of the news. The main contributions of this paper can be summarized as follows:
\begin{itemize}
\item We propose a novel deep neural network architecture for fake news detection utilizing two parallel BERT networks for headline and body of news articles.
\item We propose \textit{MaxWorth} algorithm for selecting best part of long articles for feeding into BERT network.
\item We also examined some hypotheses we had in mind to design the model. These hypotheses and their results are general and can be applied to other works related to the transfer learning in Natural Language Processing.
\end{itemize}

The paper structure is organized as follows: in section \ref{sec2}, we review some related works to our work. Section \ref{sec3} explores some concepts and models we used in our work. Section \ref{sec4}, present our proposed model and \textit{MaxWorth} algorithm that we used for text span selection. Section \ref{sec5} describes the Fakenewsnet dataset used to evaluate the model and the experiments performed on the model. Finally, in section \ref{sec6}, we conclude our discussion and suggest some plans for future research.

\section{Related Works}\label{sec2}

Approaches to fake news detection can be divided into three categories based on input: visual information-based methods, social context-based methods, and content-based methods \cite{Zhou2020}. As an example of using visual information for fake news detection we can refer to \cite{Wang2022}. In this work, Wang et al. use a multi-modal transformer using two-level visual features (MTTV) for fake news detection by images and text of news . Many works in fake news detection have been done using the context in which the news is published \cite{Kwon2013,Shu2019,Wu2015}. For example, Zhou et al. found a significant difference in the distribution of fake news hours and true news on the Sina Weibo social network during the day and night \cite{Zhou2015}. Vosoughi et al. observed that fake news is republished by Twitter users more widely than true news \cite{Vosoughi2018}. This difference can be seen in the four criteria: flow depth (number of retweet steps from the primary tweet), flow size (number of contributing users), maximum level (maximum number of contributing users per depth), and spread speed. They also found that the pattern of fake news propagation differed in different subjects; For example, they observed that fake political news spread faster on social networks than in other areas. This pattern difference also exists in different languages, websites, and topics.

\begin{figure}[t]
\centering
\includegraphics[width=0.7\textwidth]{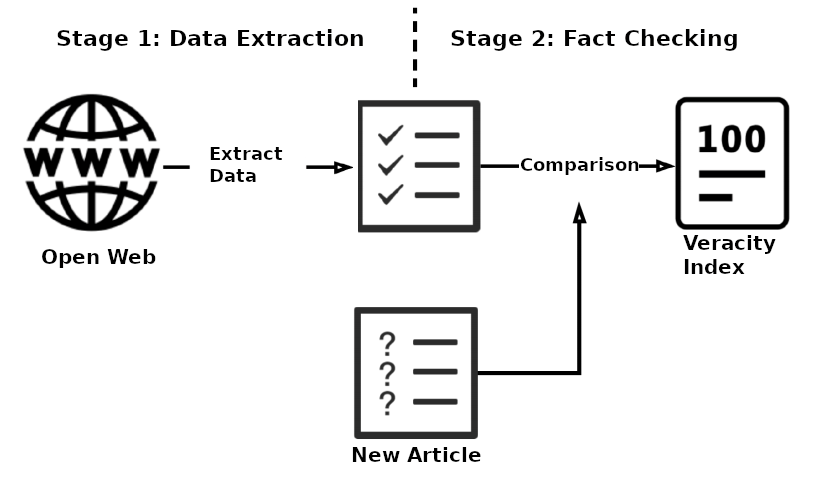}
\caption{Overall Knowledge-based Fake News Detection procedure}\label{fig2}
\end{figure}

In content-based approaches, the focus is on the text of the news. These methods are divided into two main subcategories: knowledge-based and style-based. Knowledge-based methods try to evaluate the news or claims based on fact-checking techniques. The inference of right or wrong news based on knowledge is made in two phases, and a schema can be seen in Figure \ref{fig2}. In the first phase, called knowledge extraction, knowledge is often extracted from web pages. In the second phase, the fact-checking, knowledge extracted from the new article is compared with the basic knowledge collected in the knowledge base to determine the accuracy of that news.

Thorne et al. designed a pipeline that in three phases, verifies the claim using knowledge extracted from information boxes in Wikipedia \cite{Thorne2018}. In the first phase, document retrieval, the k document closest to the claim is retrieved. In the second phase, among the k retrieved documents, a certain number of sentences most relevant to the claim are selected. In the third phase, the claim is checked against triples extracted from the chosen sentences by recognizing textual entailment \cite{Dagan2009}. Penn et al. tried identifying fake news using the knowledge graph\cite{Pan2018}. They used a knowledge graph as a valid base knowledge and constructed a TransE model of it. The TransE model is a vector representation for subject-predicate-object triples \cite{Bordes2013}. A TransE model is also built for the news article under question. Then by calculating a deviation function between two TransE models, the article is categorized into one of two categories, fake or valid. Sadeghi et al. used related and similar news published in reputable news sources as auxiliary knowledge to infer the veracity of a given news item \cite{Sadeghi2022}.

The news text analysis can take place at four levels: vocabulary, syntax, semantic, and discourse. Style-based methods try to extract some features from the text and use these features to predict the validity of the news. In \cite{Ott2011,Li2017,Abouelenien2017}, N-gram models and statistical properties of words and even letters in the text are used as features. In \cite{Ott2011,Abouelenien2017,Perez-Rosas2018}, probabilistic context-free grammar and part of Speech tags are used as features. Karimi et al. have tried to first discover the correlation between sentences as rhetorical units of the text and then use this correlation as a measure to predict news veracity \cite{Karimi2019}. Various classical learning models have been used in these studies for fake news detection, including decision trees, SVM, and KNN \cite{Mitchell2006}. Gravanis et al. conducted several experiments on various linguistic feature sets with different classifiers. They concluded that an ensemble model using ADABoost on a selected feature set, gives them the best results in detecting fake news \cite{Gravanis2019}.

However, the main challenge is to select the appropriate features of the text. There is no guarantee that the deceptions in fake news are indicated by features used in each of these models. The use of deep neural networks has the advantage that during the training process, the best features of the news can be automatically extracted. In \cite{Monti2019}, Geometric neural networks applied to the news dissemination model on Twitter have been used. Geometric neural networks derived originally from CNN, are non-Euclidean neural network used in graph and manifold data analysis \cite{Bronstein2017}.

Lang et al. have proposed an LSTM model equipped with hybrid attenuation \cite{Long2017}. They obtain the display of news text using one LSTM network and the presentation of the author's profile with another LSTM network. They categorize the news into the desired category by applying a SoftMax activation layer to the concatenated output of the two LSTM layers. Roy et al. have used an ensemble model consisting of a CNN network and a BiLSTM network \cite{Roy2019}. The results of these two models are then fed to an MLP network to apply the final classification. Thota et al. used a dense neural network on three display techniques: Bag of Words, Term Frequency–Inverse Document Frequency (TFIDF), and Word2Vec \cite{Thota2018}. They also used a dropout layer to prevent the model from over-fitting. Also Palani et al. proposed a multimodal framework that uses a BERT network for text encoding and a capsule neural network for image encoding \cite{Palani2022}.

\section{Models}\label{sec3}

In this section, we will introduce models and concepts used in our paper.

\subsection{Transfer Learning}\label{subsec31}

There is one fundamental challenge to using deep neural networks on fake news: Deep learning requires a vast amount of data to train and learn, but our access to fake news is limited. A look at the fake news datasets created so far shows that most of them have the size of a few hundred or a few thousand, which is not suitable for practicing a complex neural network with a large number of parameters \cite{Pierri2019,Wang2017}. Recent researches have used transfer learning to address this problem. In transfer learning, a pre-trained model is built by training the neural network model on a massive amount of general data that, in the case of NLP, is usually unstructured unlabeled text from the internet. Then the pre-trained model is fine-tuned on labeled data related to a specific task that shares some knowledge with the primary model \cite{Pan2009}. This learning method has been common in image-processing procedures for many years and has recently been considered in NLP.

For transfer learning in NLP, the pre-training process is usually used to build a language model. Language models generally aim to predict the next word or sentence in the text. Language models have two essential properties that make them a good choice for use as a base model. First, they can be trained on unlabeled data, and therefore have a huge amount of data available for solid learning. Second, a well-trained language model can give us a good understanding of the structure and semantics of the language, which is much needed and helpful in downstream tasks.

The language model created in this manner, replaces static word embeddings such as Word2Vec and GloVE \cite{Mikolov2013,Pennington2014}. It can be called a kind of dynamic word embedding. Static word embeddings were context independent; Although a word can give different meanings in different contexts, the word was always mapped to a fixed vector. For example, the word running in the two sentences \textit{"Robert is running a club"} and \textit{"Robert is running in a marathon"} have two completely different meanings, but the display of both of them in word2vec is the same. In transfer learning language models, the whole text is mapped to a sequence of embedding vectors, which means each word displayed by a vector according to its context.

For transfer learning in NLP, many architectures are used. For example, ELMo used a Bidirectional LSTM \cite{Peters2020}. But the most popular architecture is transformers. GPT2 and BERT are two recent states of art models that used transformers for transfer learning in NLP \cite{Devlin2019,Peters2020}.

\subsection{Transformers}\label{sec32}

Until recent years, most learning systems used gated RNNs such as LSTM and GRU to process natural language, which designed to processing sequential data. After introducing the attention mechanism, these architectures also added attention layers to their architecture to better learn the connection between the sequence components and weigh the components according to their importance in the model \cite{Bahdanau2015,Luong2015}. In recent years, in most researches, transformers have replaced the previous architectures due to the excellent results of transformers in NLP tasks. Transformers are built entirely upon the attention mechanism because there is the fact that attention alone will be able to meet the learning performance expected from the RNN \cite{Vaswani2017}.

\subsubsection{BERT}\label{321}

One of the latest and most promising transformer models is the BERT \cite{Devlin2019}. BERT, (developed by Google experts) is pre-trained for two pre-training purposes based on a large amount of unlabeled data:
\begin{itemize}
\item Masked language modeling: For this purpose, about 15\% of the words in the sentence are masked, and the model should predict them using the context of the words. It should be noted that unlike previous Transformer models such as the GPT, BERT is bidirectional, meaning that it uses pre- and post-word content to predict. It is shown that this can increase the efficiency of the model compared to previous one-directional models such as GPT in sentence-level tasks such as question-answer (SQuAD 1.1 \cite{Rajpurkar2016}) and Multi-Genre Natural Language Inference (MNLI \cite{Williams2018}). It also eliminates the dependence of the model on the direction of the language. It is just as effective for right-to-left languages as it is for left-to-right languages.
\item Next sentence prediction: By giving two sentences, the model must answer that: Is the second sentence a sub-sentence of the first sentence or not?
\end{itemize}

\begin{figure}[t]
\centering
\includegraphics[width=0.9\textwidth]{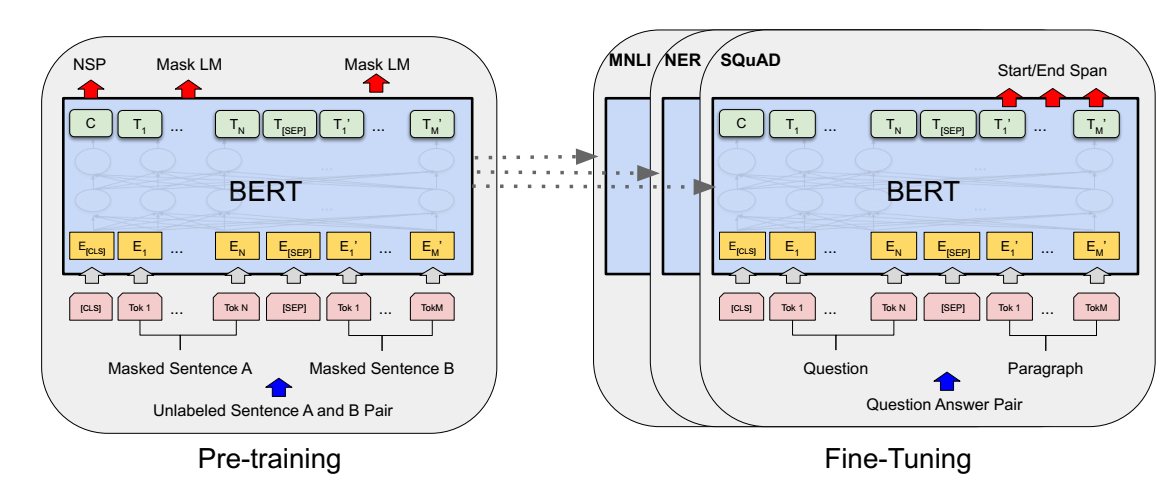}
\caption{Overall pre-training and fine-tuning procedures for BERT. Apart from output layers, the same architectures are used in pre-training and fine-tuning. The same pre-trained model parameters are used to initialize models for the different downstream tasks. Only parameters are fine-tuned \cite{Devlin2019}}
\label{fig3}
\end{figure}

The Overall pre-training and fine-tuning procedures for BERT are shown in Figure \ref{fig3}.Using BERT in downstream tasks is enough to fine-tune the BERT model by practicing it on task-specific data. In the downstream tasks, the same BERT architecture and weight are used. We only add the output layers according to the desired output of the task. When fine-tuning the model, we start with the model parameters obtained from the pre-training, and all model parameters are fine-tuned during the fine-tuning phase.

\section{Proposed Model}\label{sec4}

\begin{figure}[t]
\centering
\includegraphics[width=0.9\textwidth]{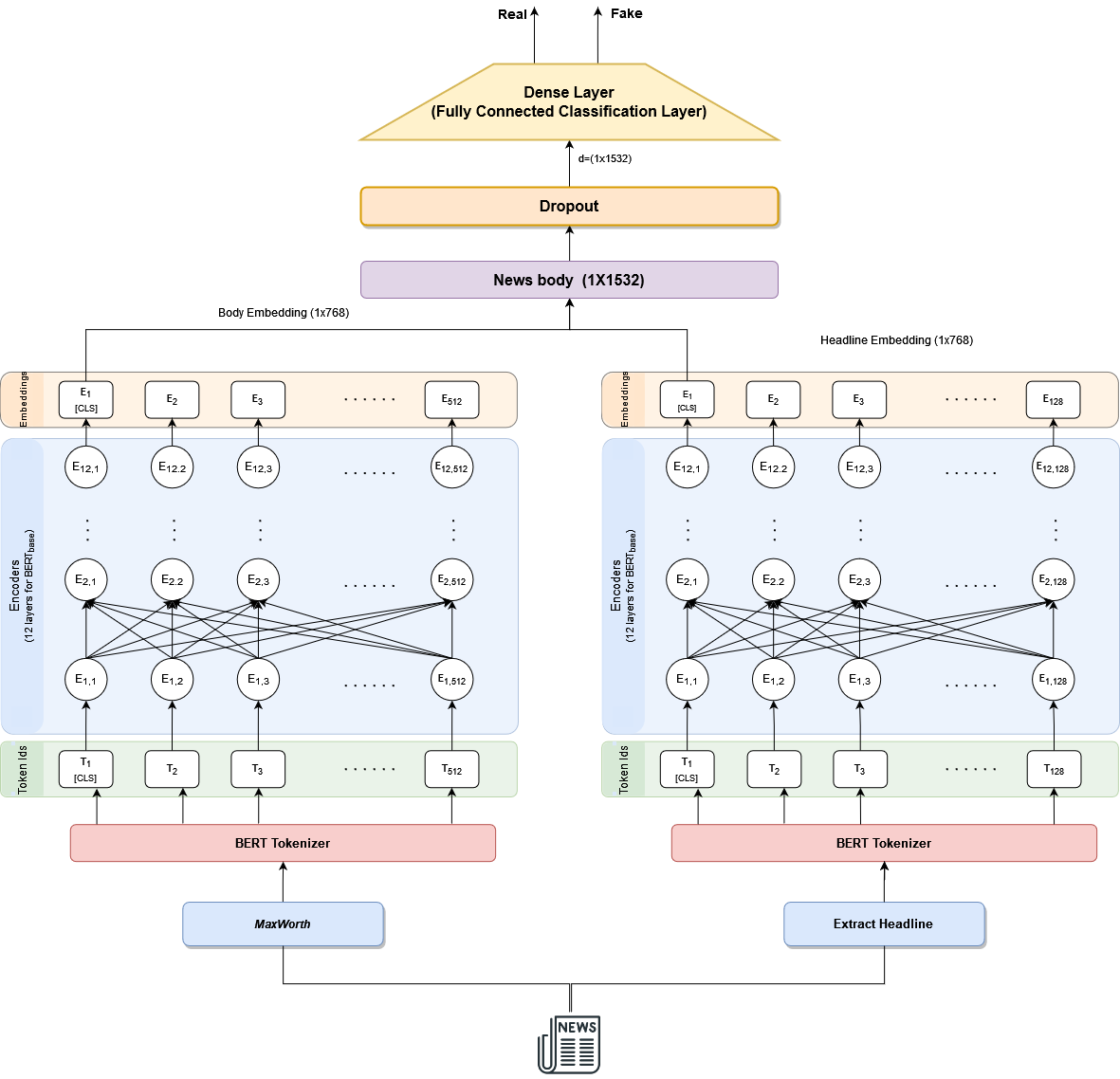}
\caption{Architecture of \textit{MWPBert} ([CLS] is a special token that is inserted at the beginning of text; encoding of whole input text calculated in [CLS] token embedding.)}
\label{fig4}
\end{figure}

A schema of our proposed model is shown in \ref{fig4}. We call it \textbf{MWPBert}, which stands for \textbf{M}ax \textbf{W}orth \textbf{P}arallel \textbf{BERT}. First, the news headline is separated from the beginning of the news with a maximum length of 128 tokens and enters into a BERT network. On the other hand, using the \textit{MaxWorth} algorithm, the most appropriate news text span is selected from the rest of the news text. The selected span text also enters into a separate BERT network with a maximum input length of 512. The output of both networks, which represents the semantics of the headline and news text, join together and feed into a dropout layer. Then a linear neural network delivers the desired output for the decision about whether the news is true or false. It should be noted that the learning structure and semantics of news text occures within encoder layers of BERT, and the output network only adapts the BERT output to the desired output. There is no need for a complex network after BERT; adding more complexity increases the time and computational costs and reduces the impact of the effective attention mechanism in the BERT encoder layers \cite{Devlin2019}. The input and output sizes of each network layer can be seen in Table \ref{tab1}.

\begin{table}\centering
\caption{Input and output size of proposed model layers}\label{tab1}
\scriptsize
\begin{tabular}{lrrr}\toprule
\textbf{Layer} &\textbf{Input size} &\textbf{Output size} \\\midrule
\textbf{BERT (text span)} &512 &768 \\
\textbf{BERT (headline)} &128 &768 \\
\textbf{Concat} &2$\times$768 &1532 \\
\textbf{Dropout} &1532 &1532 \\
\textbf{Linear} &1532 &2 \\
\bottomrule
\end{tabular}
\end{table}

\subsection{4.1	Selecting Text Span with Maximum Check-worthiness}\label{sec41}

To feed the news text, which is usually long, to BERT, whose input size is fixed and limited (this size is a maximum of 512 in the pre-trained model BERT\textsubscript{base}), we have two options; first, to increase the size of the network; Second, shorten the length of the input. The first solution is not suitable for two reasons. First, the upper limit of the news text size is unknown; and, more importantly, the bigger model has more parameters and hence needs more data to train, which in the case of fake news is not available. In addition, the time complexity of BERT network training increases with increasing input size. Therefore, we looked for a way to reduce the input size. Using text summarization algorithms is not very helpful here. Because these algorithms generally paste some sentences together that are separate and possibility unrelated to summarize the text. This dramatically reduces the structural and semantical connections between the obtained sentences. As a result, the BERT network will not be able to discover the proper structure and meaning of the primary news. In addition, the sentences that characterize the correctness or incorrectness of the text are some Check-worthy factual Sentences, not necessarily the sentences that appear in summary. Therefore, we use the \textit{MaxWorth} algorithm to select the most valuable news span for validation, instead of traditional summarizing. This algorithm is shown in algorithm \ref{algo1}.

\begin{algorithm}
\caption{\textit{MaxWorth} Algorithm for selecting news text span with maximum worth for fact-checking}
\label{algo1}
\begin{algorithmic}[1]
\Function{MaxWorth}{$text,max\_length$}
    \State $sentences \leftarrow extract\_sentences(text)$
    \State $start \leftarrow 0$
    \State $max\_score \leftarrow 0$
    \While{not reached to the last sentence}
        \State $span \leftarrow$ longest range of $sentences$ beginning at $start$ and $length<max\_len$
        \State $score \leftarrow AVERAGE(score(s)$ for $s$ IN $span)$
        \If{$score > max\_score$}
            \State $max\_score \leftarrow score$
            \State $max\_span \leftarrow span$
        \EndIf
        \State $start \leftarrow start + 1$
    \EndWhile
\State \Return $max\_span$
\EndFunction
\end{algorithmic}
\end{algorithm}
`	
The \textit{MaxWorth} algorithm uses check-worthiness sentence ranking to evaluate news text sentences. It uses ClaimBuster API for scoring each sentence \cite{Hassan2017}. ClaimBuster is a Check-worthiness sentence ranking algorithm that is used as the first step of the fact-checking process and scores news sentences based on how valuable they are for fact-checking. The basis of ClaimBuster claim spotting API is a classification algorithm that classifies sentences into three categories:
\begin{itemize}
\item \textbf{Non-Factual Sentence (NFS):} Subjective sentences and many questions are categorized as NFS. These sentences do not contain any factual claim. Here are two examples:
\begin{itemize}
\item "But I think it’s time to talk about the future."
\item "Do you remember the last time you said that?"
\end{itemize}

\item Unimportant Factual Sentence (UFS): These are factual claims but not check-worthy. The general public will not be interested in knowing whether these sentences are true or false. And fact-checkers do not find these sentences necessary for checking. Some examples
are as follows:
\begin{itemize}
\item "Next Tuesday is Election day."
\item "Two days ago we ate lunch at a restaurant."
\end{itemize}

\item Check-worthy Factual Sentence (CFS): They contain factual claims, and the general public will be interested in knowing whether the claims are true. Journalists look for these types of claims for fact-checking. Some examples are:
\begin{itemize}
\item "He voted against the first Gulf War."
\item "Over a million and a quarter of Americans are HIV-positive."
\end{itemize}
\end{itemize}

Claimbuster used 6615 features, including words, POS, entity types, and sentence length in an SVM classifier to classify sentences. The score of a sentence is considered the probability that the sentence belongs to the CFS class:
\begin{equation}
Score(x)=P(class=CFS \mid x)
\end{equation}

Since we don't want to choose separate sentences of the text to avoid losing the semantic integrity, \textit{MaxWorth} scans the news text and finds a continuous span of the text with a maximum length of 512 (BERT\textsubscript{base} model input size that we used) so that the average check-worthiness score becomes maximum.

\subsubsection{Pretrained BERT Model}
There are several pre-trained models of BERT, the most important of which you can see in Table \ref{tab2}. The evaluations of these models show that uncased models work slightly better than case. Due to the relatively small database size we use, the large model did not yield good results in our experiments, as mentioned in \cite{Devlin2019}. Therefore, in our model, we have used the uncased base model.

\begin{adjustwidth}{0 cm}{0 cm}\centering\begin{threeparttable}[!htb]
\caption{BERT's main pre-trained models}\label{tab2}
\begin{tabularx}{\textwidth}{p{2.5cm}XXXp{2.5cm}}
\toprule
\textbf{Model} &\textbf{Number of Layers} &\textbf{Hidden states} &\textbf{Cased / uncased} &\textbf{Number of parameters} \\\midrule
\textbf{BERT\textsubscript{large}} &24 &1024 &Both exist &340M \\
\textbf{BERT\textsubscript{base}} &12 &768 &Both exist &110M \\
\textbf{BERT\textsubscript{base}}-Multilingual &12 &768 &Cased &110M \\
\bottomrule
\end{tabularx}
\end{threeparttable}\end{adjustwidth}

\subsubsection{Output Network}

For token-level tasks, such as tagging a sequence or question answering, the display of individual tokens enters the output network. But for classification, the only representation of the special token at the beginning of the text ([CLS]) enters the output network representing the entire text. The output network consists of only a concatenation, a dropout layer to prevent the network from being overfitted, and a linear neural network. Adding more complexity to the output not only does not increase the efficiency of the network, but also eliminates the effect of the self-attention mechanism in BERT. The experiments in the next section show the validity of this claim.

\section{Experiments}\label{sec5}

Since we wanted to evaluate the effectiveness of the proposed model in detecting full-length fake news, we chose the Fakenewsnet dataset to test the model \cite{Shu2020}. \textit{Fakenewsnet} is one of the largest full-text news datasets available until now, containing 22,616 complete news articles collected from \textit{politifact.com} and \textit{gossipcop.com} sites. Most fake news datasets are collected from tweets, headlines, comments, and claims that have small lengths and essentially are not news. But the websites used in this dataset contain full-text news articles that sometimes last several pages. In addition, the news in this dataset has good quality and is the real news that has been noticed and spread by many users sometimes. Even the true news articles in the dataset have been suspected of being fake, which is why they have been cited on the gossipCop and Politifact sites, which only review controversial news. This is why this dataset can be categorized as a hard-to-check dataset. While some solutions achieve 99\% accuracy in some datasets, they hardly reach 85\% in this Fakenewsnet. Some statistics of the dataset can be seen in Table 3.

\begin{table}[!htp]\centering
\caption{News statistics in the Fakenewsnet dataset}\label{tab3}
\scriptsize
\begin{tabular}{lrrr}\toprule
\textbf{Source} &\textbf{Fake} &\textbf{Real} \\\midrule
\textbf{PolitiFact.com} &420 &528 \\
\textbf{GossipCop.com} &4974 &16694 \\
\bottomrule
\end{tabular}
\end{table}

The ultimate goal of any classifier is to achieve the highest possible accuracy, which is the ratio of the number of correct guesses to the total number of samples. Nevertheless, because accuracy alone can mislead us, especially in unbalanced classes, we have also measured each model's accuracy, recall, and F1-scores. These four measures are formulated as follows:

\begin{equation}
Accuracy = \frac{\abs{TP} + \abs{TN}}{\abs{TP} + \abs{TN}+ \abs{FP} + \abs{FN}}
\end{equation}
\begin{equation}
Precission = \frac{\abs{TP}}{\abs{TP} + \abs{FP}}
\end{equation}
\begin{equation}
Recall = \frac{\abs{TP}}{\abs{TP} + \abs{FN}}
\end{equation}
\begin{equation}
F1\_Score = 2 \times \frac{Precission \times Recall}{Precission + Recall}
\end{equation}

Another reliable measure that shows the learning rate of the model is the \textit{Receiver Operating Curve} (ROC), which shows how the \textit{True Positive Rate} (TPR) and \textit{False Positive Rate} (FPR) are changed with changes in the threshold of a classifier. We calculated \textit{Area Under this Curve} (AUC) as a reliable criterion for all models. We divided the dataset into 80\% for training and the remaining for testing. Then we trained models on the train part and calculated the performance mentioned above measure of the trained model on the test set.

\subsection{Evaluation of Model}\label{sec51}
We compared our proposed model with three states of art models that achieved good results in their evaluations. The first model is the three-level hierarchical attention network (3HAN), one of the most prominent recent works that use static word embedding \cite{Singhania2017}. This model uses GloVE as the embedding Layer and three layers of bidirectional GRU with the attention as word encoder, sentence encoder, and headline encoder, respectively. The second model is a basic CNN model with GloVE embedding from the main article of Fakenewsnet \cite{Shu2020}.

Another model is the FakeBERT proposed by Kaliyar et al. \cite{Kaliyar2021}. In this model, which uses the BERT embedding layer, the output layer includes several layers, including three parallel convolution layers, two Max Pooling layers, and several other regulating layers.

\begin{figure}[t]
\centering
\includegraphics[width=0.7\textwidth]{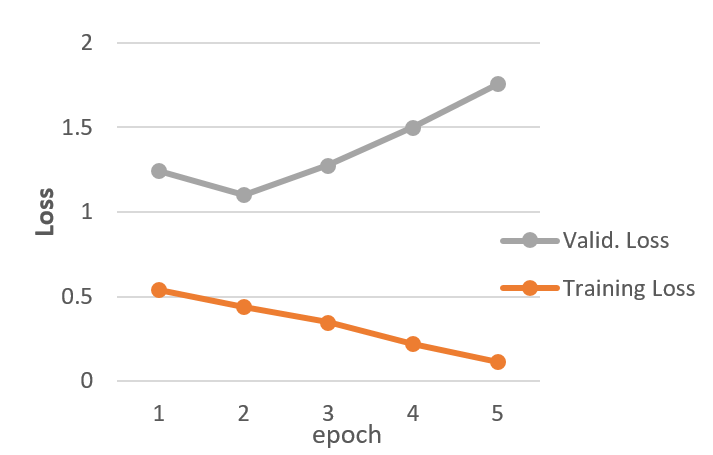}
\caption{Loss changes according to the number of epochs. As mentioned in \cite{Devlin2019}, there is no need for many training epochs through a fine-tuning phase.}
\label{fig5}
\end{figure}

Hyperparameters of training can be seen in table \ref{tab4}. Figure \ref{fig5} shows training loss and validation loss changes according to the number of epochs. As we can see, due to the pre-training of the BERT, there is no need for many training epochs to fine-tune the model like classic deep neural networks. After the second epoch model begins to overfit on training data. This is consistent with what was said in \cite{Devlin2019}.

\begin{table}[!htp]\centering
\caption{Training Hyperparameters}\label{tab4}
\scriptsize
\begin{tabular}{lrr}\toprule
\textbf{Parameter} &\textbf{Value} \\\midrule
\textbf{Learning rate} & $2\times e^{-5}$ \\
\textbf{Epoch size} &8 \\
\textbf{Optimizer} &AdamW \\
\textbf{Epsilon(of optimizer)} &$1 \times e^{-5}$ \\
\textbf{Dropout probability} & 0.2 \\
\textbf{Loss Function} &Cross Entropy \\
\bottomrule
\end{tabular}
\end{table}

\begin{figure}[t]
\centering
\includegraphics[width=0.8\textwidth]{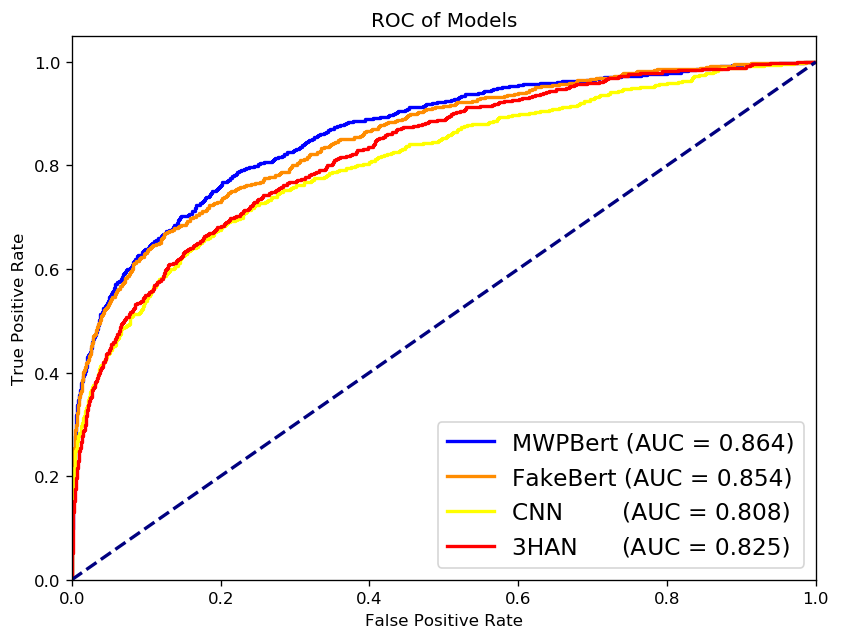}
\caption{Reciever Operation Curve of our model MWPBert and other models}
\label{fig6}
\end{figure}

\begin{figure}[t]
\centering
\includegraphics[width=\textwidth]{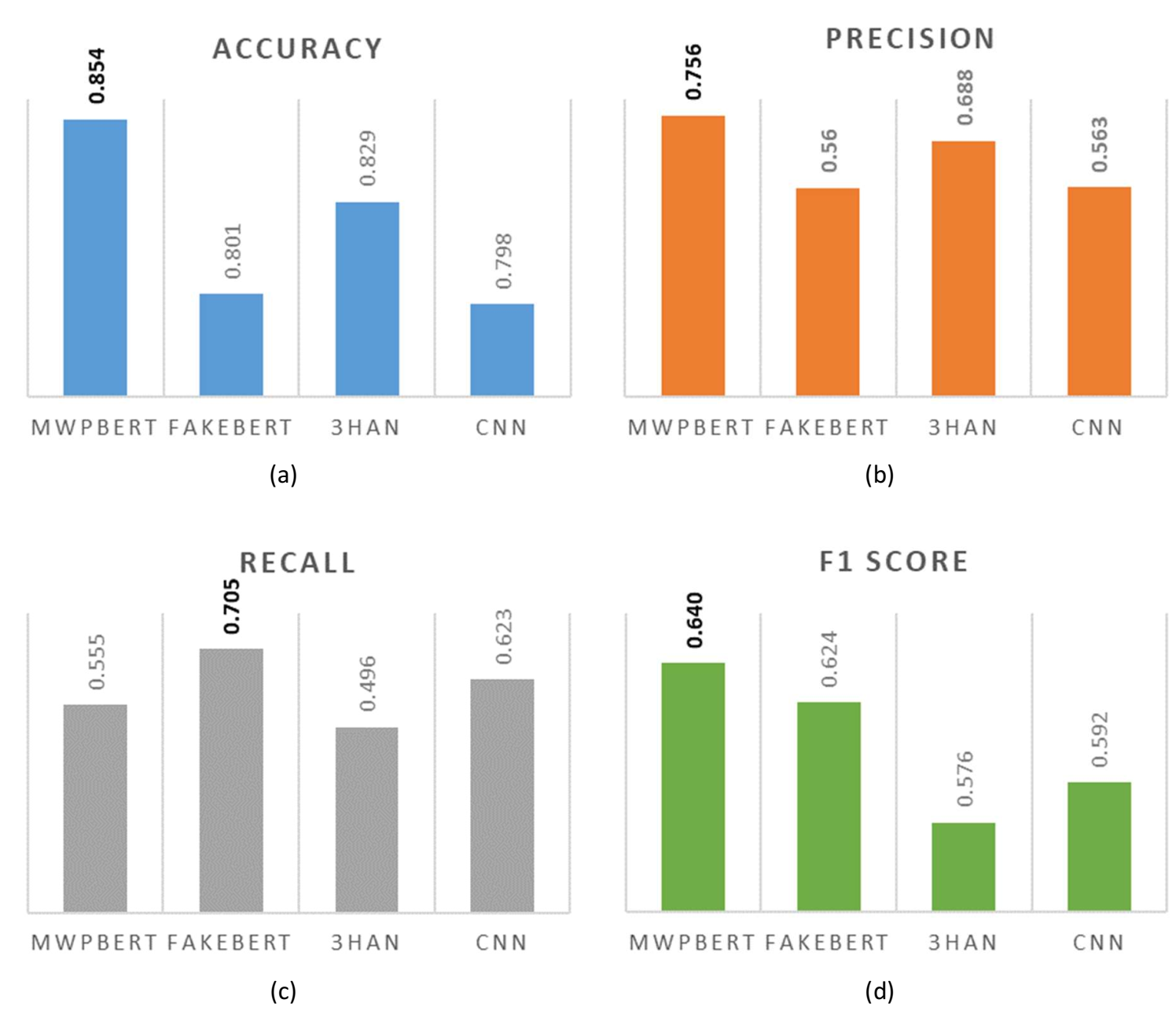}
\caption{Performance measures of models: (a): accuracy; (b): Precision, (c): Recall; (d): F1\_Score}
\label{fig7}
\end{figure}

The evaluation Results of the models can be seen in figure \ref{fig6} and figure \ref{fig7}. As we can see in figure \ref{fig6}, our proposed model, MWPBert, occupied more Area Under the Curve in the ROC of models, though the difference between MWPBert and FakeBert is too little due to their similar base language model.

\subsection{Evaluation of Hypotheses}
In addition to comparing our proposed model to other solutions, we wanted to answer some questions about the hypotheses we had in mind in designing our model. These questions were as follows:
\begin{itemize}
\item Does the use of two parallel BERT improve the performance of the model?
\item Is it reasonable to use the \textit{MaxWorth} algorithm instead of traditional text summarization methods?
\item Does the complexity of the output network affect model performance?
\item Does input text length affect model performance?
\end{itemize}
To answer these questions, we conducted some experiments, the result of which are presented in the rest of this section.
\subsubsection{Parallelization of two BERT}\label{521}
In the proposed model, we said that the two BERT networks encode the headline and the news summary in parallel, and their output is concatenated for classification. To understand the positive effect of this parallelism, it is enough to compare the results of the proposed model with the MWBERT model, which use only one BERT to encode the first 512 words of the headline and selected text span together. The results of this comparison are shown in Table \ref{tab5}. As can be seen, the performance results of MWPBert with parallelization are better than the single BERT model, indicating that using two parallel BERT has been helpful. It should be noted that although in the non-parallel mode, the precision of the model is slightly better than in the parallel mode, but its recall has dropped drastically; it can be said simply that the non-parallel model tends to categorize most of the suspicious news under the real news category.

\begin{table}[!htp]\centering
\caption{Effect of parallelization on performance}\label{tab5}
\scriptsize
\begin{tabularx}{\textwidth}{p{.3\textwidth}p{.11\textwidth}p{.11\textwidth}p{.11\textwidth}p{.11\textwidth}X}\toprule
\textbf{Model} &\textbf{Accuracy} &\textbf{Precision} &\textbf{Recall} &\textbf{F1 Score} &\textbf{RUC} \\ \midrule
\textbf{MWBERT} \newline (One Bert) &0.847 &\textbf{0.846} &0.422 &0.563 &0.848 \\
\textbf{MWPBert} \newline (2 parallel BERT) &\textbf{0.854} &0.756 &\textbf{0.555} &\textbf{0.640} &\textbf{0.863} \\
\bottomrule
\end{tabularx}
\end{table}

\subsubsection{Span Selection Method}\label{522}

In our proposed model, we said that for selecting the appropriate part of news text for feeding into BERT, instead of the traditional summarization methods, we use the \textit{MaxWorth} algorithm and find the piece of text with the most value to be checked. To verify this choice, we once summarized the text using TFIDF, used that summary, and once again fed the first 512 tokens of text into BERT without any summarization. The results of these experiments are shown in table \ref{tab6}. As can be seen, \textit{MaxWorth} has done slightly better than others in this subject.

\begin{table}[!htp]\centering
\caption{Effect of text span selection method on performance}\label{tab6}
\scriptsize
\begin{tabularx}{\textwidth}{p{.3\textwidth}p{.11\textwidth}p{.11\textwidth}p{.11\textwidth}p{.11\textwidth}X}\toprule
\textbf{Selection method} &\textbf{Accuracy} &\textbf{Precision} &\textbf{Recall} &\textbf{F1 Score} &\textbf{RUC} \\\midrule
\textbf{MaxWorth} &\textbf{0.854} &0.756 &0.555 &\textbf{0.640} &\textbf{0.863} \\
\textbf{TFIDF} &0.810 &0.579 &\textbf{0.693} &0.631 &0.849 \\
\textbf{First 512 word} &0.847 &\textbf{0.846} &0.422 &0.560 &0.848 \\
\bottomrule
\end{tabularx}
\end{table}

\subsubsection{Output Network Architecture}\label{523}

In Section \ref{sec4}, we argued that the architecture of the output network should be as simple as possible so that the BERT Self-attention mechanism can work well. We have compared our 3 BERT-based models of different output networks to investigate this claim. As the simple output network model, we used a model without check-worthy text selection, having only one BERT layer for headline and text, followed by a dropout and a linear neural network. The first complex model is the same FakeBERT model that we introduced before. As mentioned earlier, FakeBERT has a complex output network, including three convolutional neural networks and many other helper layers. In another architecture design, we added two LSTM layers and two layers of dropout to the output network of the simple output model mentioned above. As the results of experiments can be seen in Table \ref{tab7}, complex output networks add nothing more to these BERT-based models.

\begin{table}[!htp]\centering
\caption{Effect of output network architecture on performance}\label{tab7}
\scriptsize
\begin{tabularx}{\textwidth}{p{.3\textwidth}p{.11\textwidth}p{.11\textwidth}p{.11\textwidth}p{.11\textwidth}X}\toprule
\textbf{Output Network (Model)} &\textbf{Accuracy} &\textbf{Precision} &\textbf{Recall} &\textbf{F1 Score} &\textbf{RUC} \\\midrule
\textbf{Complex CNNs} \newline (FakeBERT) &0.801 &0.560 &\textbf{0.705} &0.624 &0.854 \\
\textbf{LSTM} \newline (MWPBert) &0.845 &0.703 &0.582 &0.637 &0.840 \\
\textbf{Linear} \newline (MWPBert) &\textbf{0.854} &\textbf{0.756} &0.555 &\textbf{0.640} &\textbf{0.863} \\
\bottomrule
\end{tabularx}
\end{table}

\subsubsection{Input Size}

Training time increases by increasing the size of the input. So we must have good reasons for increasing the input size. We decided to measure the model's performance by varying the input length in a model with the single BERT layer to see how increasing the input length would be a good option for us. Therefore, we trained the single BERT model with three different sizes of input: 128, 256, and 512. As seen in table \ref{tab8}, increasing the input size slightly improves the model's performance. However, bigger input sizes require more memory, and we can not increase input size to any desired value due to our computing facilities.

\begin{table}[!htp]\centering
\caption{Effect of input length on performance}\label{tab8}
\scriptsize
\begin{tabularx}{\textwidth}{p{.3\textwidth}p{.11\textwidth}p{.11\textwidth}p{.11\textwidth}p{.11\textwidth}X}\toprule
\textbf{Input size} &\textbf{Accuracy} &\textbf{Precision} &\textbf{Recall} &\textbf{F1 Score} &\textbf{RUC} \\\midrule
\textbf{128} &0.751 &0.480 &\textbf{0.770} &0.592 &0.850 \\
\textbf{256} &0.811 &0.580 &0.698 &\textbf{0.634} &\textbf{0.857} \\
\textbf{512} &\textbf{0.847} &\textbf{0.846} &0.422 &0.563 &0.848 \\
\bottomrule
\end{tabularx}
\end{table}

\section{Conclusion and Future Works}\label{sec6}

Fake news detection is a text classification task that, due to its nature, has unique features in addition to general texts. Among other features, news can be long, not all news items are equally crucial in veracity detection, and they have a headline. We tried to design a classifier based on these features that would work best. To do this, we extracted the headline from the news and the most valuable part of the news to check from its text using the \textit{MaxWorth} algorithm. We then encoded them using two parallel BERT networks and fed them into a linear classifier. The results showed that the choices made were the right choices.

Experiment results showed that for selecting part of long articles as network input, the use of the \textit{MaxWorth} algorithm shows better performance than the summarization algorithms. Also, using two parallel BERT networks for the headline and body of the news was the right choice and gave us better performance than the simple use of the BERT network. Our experience showed that the complex output networks after the BERT network not only increase the calculation overloads but also reduce the effect of BERT network weights and the model's efficiency. It is better to use a simple linear neural network in the output to adapt the output of the BERT to the expected output of the model. It is also seen that increasing the input length of BERT slightly increases the model's efficiency.

To get better results in fake news detection using deep neural networks, what we need most is more data. With a larger dataset, we can also use the larger transformer models, which can learn better with more parameters. Therefore, one of the most important things to do in the future is to collect as much fake news as possible. Some pre-trained models of BERT are multi-language. So we can see how much-learned weight in one language can be transferred to another. This way, not only can our fake news samples be gathered from different languages, but we will also have a tool to verify the news in low-resource languages. Another of our plans is to use other existing modalities of the news to detect whether it is fake. These modalities can be the context of news, including comments and feedback on social networks, non-textual news modalities such as photos, videos, and audio, and how news is published on social networks.

\section*{Declarations}

\begin{itemize}
\item \textbf{Data availability:} The datasets analysed during the current study are available in this repository: https://github.com/KaiDMML/FakeNewsNet
\item \textbf{Code availability:} The code are available from the corresponding author on reasonable request.
Declarations
\item \textbf{Conflict of interest:} The authors declare that they have no conflict of interest.
\end{itemize}


\bibliography{library}

\begin{thebibliography}{45}
\providecommand{\natexlab}[1]{#1}
\providecommand{\url}[1]{{#1}}
\providecommand{\urlprefix}{URL }
\providecommand{\doi}[1]{\url{https://doi.org/#1}}
\providecommand{\eprint}[2][]{\url{#2}}
 \bibcommenthead

\bibitem[{Abouelenien et~al(2017)Abouelenien, P{\'{e}}rez-Rosas, Zhao,
  Mihalcea, and Burzo}]{Abouelenien2017}
Abouelenien M, P{\'{e}}rez-Rosas V, Zhao B, et~al (2017) {Gender-based
  multimodal deception detection}. In: Proceedings of the ACM Symposium on
  Applied Computing, vol Part F1280. Association for Computing Machinery, New
  York, New York, USA, pp 137--144, \doi{10.1145/3019612.3019644},
  \urlprefix\url{http://dl.acm.org/citation.cfm?doid=3019612.3019644}

\bibitem[{Albawi et~al(2018)Albawi, Mohammed, and
  Al-Zawi}]{albawi2017understanding}
Albawi S, Mohammed TA, Al-Zawi S (2018) {Understanding of a convolutional
  neural network}. In: Proceedings of 2017 International Conference on
  Engineering and Technology, ICET 2017, Ieee, pp 1--6,
  \doi{10.1109/ICEngTechnol.2017.8308186}

\bibitem[{Bahdanau et~al(2015)Bahdanau, Cho, and Bengio}]{Bahdanau2015}
Bahdanau D, Cho KH, Bengio Y (2015) {Neural machine translation by jointly
  learning to align and translate}. In: 3rd International Conference on
  Learning Representations, ICLR 2015 - Conference Track Proceedings,
  \eprint{1409.0473}

\bibitem[{Bordes et~al(2013)Bordes, Usunier, Garcia-Dur{\'{a}}n, Weston, and
  Yakhnenko}]{Bordes2013}
Bordes A, Usunier N, Garcia-Dur{\'{a}}n A, et~al (2013) {Translating embeddings
  for modeling multi-relational data}. In: Advances in Neural Information
  Processing Systems

\bibitem[{Bronstein et~al(2017)Bronstein, Bruna, Lecun, Szlam, and
  Vandergheynst}]{Bronstein2017}
Bronstein MM, Bruna J, Lecun Y, et~al (2017) {Geometric Deep Learning: Going
  beyond Euclidean data}. \doi{10.1109/MSP.2017.2693418}, \eprint{1611.08097}

\bibitem[{Dagan et~al(2009)Dagan, Dolan, Magnini, and Roth}]{Dagan2009}
Dagan I, Dolan B, Magnini B, et~al (2009) {Recognizing textual entailment:
  Rational, evaluation and approaches}. \doi{10.1017/S1351324909990209}

\bibitem[{Devlin et~al(2019)Devlin, Chang, Lee, and Toutanova}]{Devlin2019}
Devlin J, Chang MW, Lee K, et~al (2019) {BERT: Pre-training of deep
  bidirectional transformers for language understanding}. NAACL HLT 2019 - 2019
  Conference of the North American Chapter of the Association for Computational
  Linguistics: Human Language Technologies - Proceedings of the Conference
  1(Mlm):4171--4186.
  {\href{https://arxiv.org/abs/1810.04805}{{https://arxiv.org/abs/arXiv:1810.04805}}}

\bibitem[{Gravanis et~al(2019)Gravanis, Vakali, Diamantaras, and
  Karadais}]{Gravanis2019}
Gravanis G, Vakali A, Diamantaras K, et~al (2019) {Behind the cues: A
  benchmarking study for fake news detection}. Expert Systems with Applications
  128:201--213. \doi{10.1016/j.eswa.2019.03.036}

\bibitem[{Hassan et~al(2017)Hassan, Arslan, Li, and Tremayne}]{Hassan2017}
Hassan N, Arslan F, Li C, et~al (2017) {Toward automated fact-checking:
  Detecting check-worthy factual claims by claimbuster}. Proceedings of the ACM
  SIGKDD International Conference on Knowledge Discovery and Data Mining Part
  F1296:1803--1812. \doi{10.1145/3097983.3098131}

\bibitem[{Kaliyar et~al(2021)Kaliyar, Goswami, and Narang}]{Kaliyar2021}
Kaliyar RK, Goswami A, Narang P (2021) {FakeBERT: Fake news detection in social
  media with a BERT-based deep learning approach}. Multimedia Tools and
  Applications 80(8):11,765--11,788. \doi{10.1007/s11042-020-10183-2}

\bibitem[{Karimi and Tang(2019)}]{Karimi2019}
Karimi H, Tang J (2019) {Learning hierarchical discourse-level structure for
  fake news detection}. NAACL HLT 2019 - 2019 Conference of the North American
  Chapter of the Association for Computational Linguistics: Human Language
  Technologies - Proceedings of the Conference 1:3432--3442.
  \doi{10.18653/v1/n19-1347},
  {\href{https://arxiv.org/abs/1903.07389}{{https://arxiv.org/abs/arXiv:1903.07389}}}

\bibitem[{Kay et~al(1999)Kay, Taaffe, and Marino}]{Mitchell2006}
Kay D, Taaffe DR, Marino FE (1999) {Whole-body pre-cooling and heat storage
  during self-paced cycling performance in warm humid conditions}. Journal of
  Sports Sciences 17(12):937--944. \doi{10.1080/026404199365326},
  {\href{https://arxiv.org/abs/9605103}{{https://arxiv.org/abs/arXiv:9605103}}}
  {[cs]}

\bibitem[{Kowsari et~al(2019)Kowsari, Meimandi, Heidarysafa, Mendu, Barnes, and
  Brown}]{Kowsari2019}
Kowsari K, Meimandi KJ, Heidarysafa M, et~al (2019) {Text classification
  algorithms: A survey}. Information (Switzerland) 10(4):150.
  \doi{10.3390/info10040150},
  {\href{https://arxiv.org/abs/1904.08067}{{https://arxiv.org/abs/arXiv:1904.08067}}}

\bibitem[{Kwon et~al(2013)Kwon, Cha, Jung, Chen, and Wang}]{Kwon2013}
Kwon S, Cha M, Jung K, et~al (2013) {Prominent features of rumor propagation in
  online social media}. In: Proceedings - IEEE International Conference on Data
  Mining, ICDM, pp 1103--1108, \doi{10.1109/ICDM.2013.61}

\bibitem[{Li et~al(2017)Li, Qin, Ren, and Liu}]{Li2017}
Li L, Qin B, Ren W, et~al (2017) {Document representation and feature
  combination for deceptive spam review detection}. Neurocomputing 254:33--41.
  \doi{10.1016/j.neucom.2016.10.080}

\bibitem[{Long et~al(2017)Long, Lu, Xiang, Li, and Huang}]{Long2017}
Long Y, Lu Q, Xiang R, et~al (2017) {Fake News Detection Through
  Multi-Perspective Speaker Profiles}. Proceedings of the Eighth International
  Joint Conference on Natural Language Processing Volume 2:(8):252--256.
  \urlprefix\url{http://www.aclweb.org/anthology/I17-2043}

\bibitem[{Luong et~al(2015)Luong, Pham, and Manning}]{Luong2015}
Luong MT, Pham H, Manning CD (2015) {Effective approaches to attention-based
  neural machine translation}. In: Conference Proceedings - EMNLP 2015:
  Conference on Empirical Methods in Natural Language Processing, pp
  1412--1421, \doi{10.18653/v1/d15-1166}, \eprint{1508.04025}

\bibitem[{Meel and Vishwakarma(2020)}]{Meel2020}
Meel P, Vishwakarma DK (2020) {Fake news, rumor, information pollution in
  social media and web: A contemporary survey of state-of-the-arts, challenges
  and opportunities}. \doi{10.1016/j.eswa.2019.112986}

\bibitem[{Mikolov et~al(2013)Mikolov, Sutskever, Chen, Corrado, and
  Dean}]{Mikolov2013}
Mikolov T, Sutskever I, Chen K, et~al (2013) {Distributed representations
  ofwords and phrases and their compositionality}. In: Advances in Neural
  Information Processing Systems

\bibitem[{Monti et~al(2019)Monti, Frasca, Eynard, Mannion, and
  Bronstein}]{Monti2019}
Monti F, Frasca F, Eynard D, et~al (2019) {Fake News Detection on Social Media
  using Geometric Deep Learning}. Arxiv 1902-06673v1 pp 1--15.
  \urlprefix\url{http://arxiv.org/abs/1902.06673},
  {\href{https://arxiv.org/abs/1902.06673}{{https://arxiv.org/abs/arXiv:1902.06673}}}

\bibitem[{Ott et~al(2011)Ott, Choi, Cardie, and Hancock}]{Ott2011}
Ott M, Choi Y, Cardie C, et~al (2011) {Finding deceptive opinion spam by any
  stretch of the imagination}. In: ACL-HLT 2011 - Proceedings of the 49th
  Annual Meeting of the Association for Computational Linguistics: Human
  Language Technologies, pp 309--319, \eprint{1107.4557}

\bibitem[{Palani et~al(2022)Palani, Elango, and {Vignesh
  Viswanathan}}]{Palani2022}
Palani B, Elango S, {Vignesh Viswanathan} K (2022) {CB-Fake: A multimodal deep
  learning framework for automatic fake news detection using capsule neural
  network and BERT}. Multimedia Tools and Applications 81(4):5587--5620.
  \doi{10.1007/s11042-021-11782-3},
  \urlprefix\url{https://link.springer.com/article/10.1007/s11042-021-11782-3}

\bibitem[{Pan et~al(2018)Pan, Pavlova, Li, Li, Li, and Liu}]{Pan2018}
Pan JZ, Pavlova S, Li C, et~al (2018) {Content based fake news detection using
  knowledge graphs}. In: Lecture Notes in Computer Science (including subseries
  Lecture Notes in Artificial Intelligence and Lecture Notes in
  Bioinformatics), vol 11136 LNCS. Springer, pp 669--683,
  \doi{10.1007/978-3-030-00671-6_39},
  \urlprefix\url{http://link.springer.com/10.1007/978-3-030-00671-6{\_}39}

\bibitem[{Panigrahi et~al(2021)Panigrahi, Nanda, and Swarnkar}]{Pan2009}
Panigrahi S, Nanda A, Swarnkar T (2021) {A Survey on Transfer Learning}. Smart
  Innovation, Systems and Technologies 194(10):781--789.
  \doi{10.1007/978-981-15-5971-6_83}

\bibitem[{Pennington et~al(2014)Pennington, Socher, and
  Manning}]{Pennington2014}
Pennington J, Socher R, Manning CD (2014) {GloVe: Global vectors for word
  representation}. In: EMNLP 2014 - 2014 Conference on Empirical Methods in
  Natural Language Processing, Proceedings of the Conference, pp 1532--1543,
  \doi{10.3115/v1/d14-1162}

\bibitem[{P{\'{e}}rez-Rosas et~al(2018)P{\'{e}}rez-Rosas, Kleinberg, Lefevre,
  and Mihalcea}]{Perez-Rosas2018}
P{\'{e}}rez-Rosas V, Kleinberg B, Lefevre A, et~al (2018) {Automatic detection
  of fake news}. \urlprefix\url{https://www.aclweb.org/anthology/C18-1287/},
  \eprint{1708.07104}

\bibitem[{Peters et~al(2018)Peters, Neumann, Zettlemoyer, and Yih}]{Peters2020}
Peters ME, Neumann M, Zettlemoyer L, et~al (2018) {Dissecting contextual word
  embeddings: Architecture and representation}. In: Proceedings of the 2018
  Conference on Empirical Methods in Natural Language Processing, EMNLP 2018,
  pp 1499--1509, \doi{10.18653/v1/d18-1179}, \eprint{1808.08949}

\bibitem[{Pierri and Ceri(2019)}]{Pierri2019}
Pierri F, Ceri S (2019) {False news on social media: A data-driven survey}.
  SIGMOD Record 48(2):18--32. \doi{10.1145/3377330.3377334},
  {\href{https://arxiv.org/abs/1902.07539}{{https://arxiv.org/abs/arXiv:1902.07539}}}

\bibitem[{Rajpurkar et~al(2016)Rajpurkar, Zhang, Lopyrev, and
  Liang}]{Rajpurkar2016}
Rajpurkar P, Zhang J, Lopyrev K, et~al (2016) {SQuad: 100,000+ questions for
  machine comprehension of text}. In: EMNLP 2016 - Conference on Empirical
  Methods in Natural Language Processing, Proceedings, pp 2383--2392,
  \doi{10.18653/v1/d16-1264}, \eprint{1606.05250}

\bibitem[{Roy et~al(2019)Roy, Basak, Ekbal, and Bhattacharyya}]{Roy2019}
Roy A, Basak K, Ekbal A, et~al (2019) {A Deep Ensemble Framework for Fake News
  Detection and Multi-Class Classification of Short Political Statements}. In:
  Proceedings of the 16th International Conference on Natural Language
  Processing

\bibitem[{Sadeghi et~al(2022)Sadeghi, Bidgoly, and Amirkhani}]{Sadeghi2022}
Sadeghi F, Bidgoly AJ, Amirkhani H (2022) {Fake news detection on social media
  using a natural language inference approach}. Multimedia Tools and
  Applications 81(23):33,801--33,821. \doi{10.1007/s11042-022-12428-8},
  \urlprefix\url{https://link.springer.com/article/10.1007/s11042-022-12428-8}

\bibitem[{Shu et~al(2019)Shu, Mahudeswaran, and Liu}]{Shu2019}
Shu K, Mahudeswaran D, Liu H (2019) {FakeNewsTracker: a tool for fake news
  collection, detection, and visualization}. Computational and Mathematical
  Organization Theory 25(1):60--71. \doi{10.1007/s10588-018-09280-3},
  \urlprefix\url{https://doi.org/10.1007/s10588-018-09280-3}

\bibitem[{Shu et~al(2020)Shu, Mahudeswaran, Wang, Lee, and Liu}]{Shu2020}
Shu K, Mahudeswaran D, Wang S, et~al (2020) {FakeNewsNet: A Data Repository
  with News Content, Social Context, and Spatiotemporal Information for
  Studying Fake News on Social Media}. Big Data 8(3):171--188.
  \doi{10.1089/big.2020.0062},
  {\href{https://arxiv.org/abs/1809.01286}{{https://arxiv.org/abs/arXiv:1809.01286}}}

\bibitem[{Singhania et~al(2017)Singhania, Fernandez, and Rao}]{Singhania2017}
Singhania S, Fernandez N, Rao S (2017) {3HAN: A Deep Neural Network for Fake
  News Detection}. Lecture Notes in Computer Science (including subseries
  Lecture Notes in Artificial Intelligence and Lecture Notes in Bioinformatics)
  10635 LNCS:572--581. \doi{10.1007/978-3-319-70096-0_59}

\bibitem[{Thorne and Vlachos(2018)}]{Thorne2018}
Thorne J, Vlachos A (2018) {Automated fact checking: Task formulations, methods
  and future directions}. COLING 2018 - 27th International Conference on
  Computational Linguistics, Proceedings pp 3346--3359.
  {\href{https://arxiv.org/abs/1806.07687}{{https://arxiv.org/abs/arXiv:1806.07687}}}

\bibitem[{Thota et~al(2018)Thota, Tilak, Ahluwalia, and Lohia}]{Thota2018}
Thota A, Tilak P, Ahluwalia S, et~al (2018) {Fake news detection: a deep
  learning approach}. SMU Data Science Review 1(3):10.
  \urlprefix\url{https://scholar.smu.edu/datasciencereview/vol1/iss3/10}

\bibitem[{Vaswani et~al(2017)Vaswani, Shazeer, Parmar, Uszkoreit, Jones, Gomez,
  Kaiser, and Polosukhin}]{Vaswani2017}
Vaswani A, Shazeer N, Parmar N, et~al (2017) {Attention is all you need}. In:
  Advances in Neural Information Processing Systems, pp 5999--6009,
  \eprint{1706.03762}

\bibitem[{Vosoughi et~al(2018)Vosoughi, Roy, and Aral}]{Vosoughi2018}
Vosoughi S, Roy D, Aral S (2018) {The spread of true and false news online}.
  Science 359(6380):1146--1151. \doi{10.1126/science.aap9559}

\bibitem[{Wang et~al(2022)Wang, Feng, cai Xiong, heng Wang, and hua
  Qiang}]{Wang2022}
Wang B, Feng Y, cai Xiong X, et~al (2022) {Multi-modal transformer using
  two-level visual features for fake news detection}. Applied Intelligence pp
  1--15. \doi{10.1007/s10489-022-04055-5},
  \urlprefix\url{https://link.springer.com/article/10.1007/s10489-022-04055-5}

\bibitem[{Wang(2017)}]{Wang2017}
Wang WY (2017) {“Liar, liar pants on fire”: A new benchmark dataset for
  fake news detection}. In: ACL 2017 - 55th Annual Meeting of the Association
  for Computational Linguistics, Proceedings of the Conference (Long Papers),
  pp 422--426, \doi{10.18653/v1/P17-2067}, \eprint{1705.00648}

\bibitem[{Williams et~al(2018)Williams, Nangia, and Bowman}]{Williams2018}
Williams A, Nangia N, Bowman SR (2018) {A broad-coverage challenge corpus for
  sentence understanding through inference}. In: NAACL HLT 2018 - 2018
  Conference of the North American Chapter of the Association for Computational
  Linguistics: Human Language Technologies - Proceedings of the Conference, pp
  1112--1122, \doi{10.18653/v1/n18-1101}, \eprint{1704.05426}

\bibitem[{Wolf et~al(2020)Wolf, Debut, Sanh, Chaumond, Delangue, Moi, Cistac,
  Rault, Louf, Funtowicz, Davison, Shleifer, von Platen, Ma, Jernite, Plu, Xu,
  {Le Scao}, Gugger, Drame, Lhoest, and Rush}]{Wolf2020}
Wolf T, Debut L, Sanh V, et~al (2020) {Transformers: State-of-the-Art Natural
  Language Processing}. In: Proceedings of the 2020 Conference on Empirical
  Methods in Natural Language Processing: System Demonstrations, pp 38--45,
  \doi{10.18653/v1/2020.emnlp-demos.6}

\bibitem[{Wu et~al(2015)Wu, Yang, and Zhu}]{Wu2015}
Wu K, Yang S, Zhu KQ (2015) {False rumors detection on Sina Weibo by
  propagation structures}. In: Proceedings - International Conference on Data
  Engineering, pp 651--662, \doi{10.1109/ICDE.2015.7113322}

\bibitem[{Zhou and Zafarani(2020)}]{Zhou2020}
Zhou X, Zafarani R (2020) {A Survey of Fake News: Fundamental Theories,
  Detection Methods, and Opportunities}. ACM Computing Surveys 53(5).
  \doi{10.1145/3395046},
  {\href{https://arxiv.org/abs/1812.00315}{{https://arxiv.org/abs/arXiv:1812.00315}}}

\bibitem[{Zhou et~al(2015)Zhou, Cao, Jin, Xie, Su, Zhang, Chu, and
  Cao}]{Zhou2015}
Zhou X, Cao J, Jin Z, et~al (2015) {Real-time news certification system on sina
  weibo}. In: WWW 2015 Companion - Proceedings of the 24th International
  Conference on World Wide Web, pp 983--988, \doi{10.1145/2740908.2742571}

\end{thebibliography}


\end{document}